\documentclass[runningheads]{llncs}

\usepackage[mobile]{eccv}

\usepackage{eccvabbrv}

\usepackage{graphicx}
\usepackage{booktabs}
\usepackage{amsmath}
\usepackage{bbding}
\usepackage{amssymb}
\usepackage[dvipsnames]{xcolor}

\usepackage[accsupp]{axessibility}  
\usepackage{bm}
\usepackage{amssymb}
\usepackage{makecell}

\newcommand\blfootnote[1]{%
  \begingroup
  \renewcommand\thefootnote{}\footnote{#1}%
  \addtocounter{footnote}{-1}%
  \endgroup
}

\usepackage[pagebackref,breaklinks,colorlinks,citecolor=eccvblue]{hyperref}

\usepackage{orcidlink}

\begin{document}

\title{\textcolor{RoyalPurple}{CondSeg}: Ellipse Estimation of Pupil and Iris via \textcolor{RoyalPurple}{Cond}itioned \textcolor{RoyalPurple}{Seg}mentation} 


\author{Zhuang Jia$^*$ \and
Jiangfan Deng$^*$ \and
Liying Chi$^*$ \and 
Xiang Long \and 
Daniel K. Du
}

\authorrunning{Z. Jia et al.}

\institute{Bytedance Inc.
}

\maketitle

\blfootnote{
\noindent $*$ Equal contribution. \\
}

\vspace{-25pt}

\begin{abstract}
    Parsing of eye components (i.e. pupil, iris and sclera) is fundamental for eye tracking and gaze estimation for AR/VR products. Mainstream approaches tackle this problem as a multi-class segmentation task, providing only visible part of pupil/iris, other methods regress elliptical parameters using human-annotated full pupil/iris parameters. In this paper, we consider two priors: projected full pupil/iris circle can be modelled with ellipses (\textbf{ellipse prior}), and the visibility of pupil/iris is controlled by openness of eye-region (\textbf{condition prior}), and design a novel method \textbf{CondSeg} to estimate elliptical parameters of pupil/iris directly from segmentation labels, without explicitly annotating full ellipses, and use eye-region mask to control the visibility of estimated pupil/iris ellipses. Conditioned segmentation loss is used to optimize the parameters by transforming parameterized ellipses into pixel-wise soft masks in a differentiable way. Our method is tested on public datasets (OpenEDS-2019/-2020) and shows competitive results on segmentation metrics, and provides accurate elliptical parameters for further applications of eye tracking simultaneously.
\end{abstract}

\section{Introduction}
\label{sec:intro}

Obtaining precise gaze estimation or eye tracking (ET) is of great importance in many areas, including the currently popular AR (augmented reality) and VR (virtual reality) applications. In AR/VR products, the estimated gaze is utilized for foveated rendering, user interaction and other tasks. A fundamental necessity for calculating the gaze direction is to identify the locations and contours of the components in the eye image, i.e. the pupil, iris and sclera\cite{guestrin2006general, sigut2010iris}. This task is commonly recognized as the multi-class segmentation task, which can be solved using the learning-based segmentation approaches. Therefore, various methods are proposed to optimize the segmentation results by designing proper network architectures and augmentation strategies\cite{chaudhary2019ritnet, alkassar2015robust, chaudhary2021semi, park2018learning}. Since the need for pupil and iris is their elliptical information to estimate eye model parameters or gaze direction, the segmented mask is then fitted to elliptical parameters to generate the final result.

Another way for estimating the full ellipse of pupil or iris is to use a trainable network to directly predict the full mask using the eye image, or to regress the elliptical parameters (generally denoted by the 5D vector $(x_0,y_0,a,b,\theta)$, indicating the center coordinate, semi-major/semi-minor axis length, and the direction angle). It is more concise in estimating the bio-metric features of eye by this means, but the drawback is that labeling the ellipses of full pupil/iris is required. Labeling ellipse by changing its position, shape and rotation is quite laborious, as the annotators need to carefully match the ellipse boundary to pupil/iris edges in the visible region, while also keep the shape and rotation of ellipse reasonable \cite{fuhl2021teyed}. Empirically, the annotation of full pupil/iris in ellipse format is about $2\times \sim 3\times$ in time consumption compared with common segmentation mask format. This obstacle makes it advantageous to design a more elegant pipeline to estimate the elliptical parameters without explicitly labeling them.

\begin{figure}[ht]
  \centering
  \includegraphics[height=2.5cm]{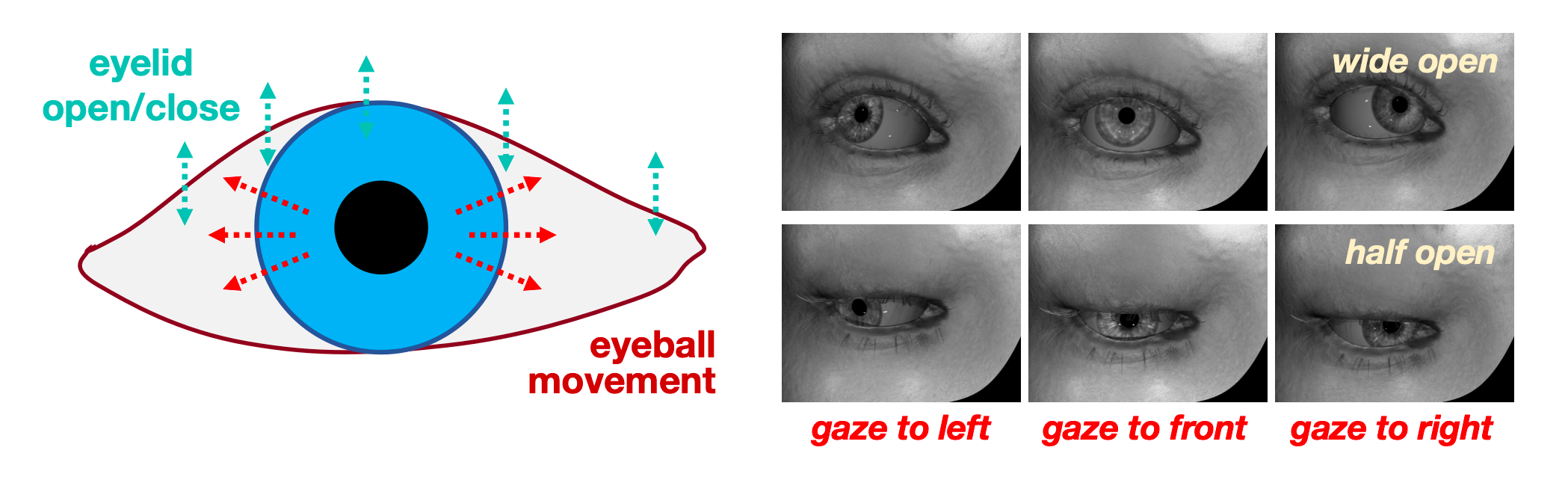}
  \caption{eye-region appearance in common eye images can be decoupled in two dimensions: the iris/pupil position which is controlled by eyeball movement and gaze direction, and the eyelid openness which is related to the elevation of upper eyelid controlled by voluntary muscle (i.e. levator palpebrae superioris). (Synthetic images on the right are from NVGaze dataset \cite{kim2019nvgaze})}
  \label{fig:decouple}
\end{figure}

In this paper, we deal with pupil and iris parsing as a conditioned segmentation task, which allows us to estimate the full pupil/iris region using visible-only annotations (only the visible part of pupil/iris mask is accessible). 
This idea is based on the prior that the pupil/iris segmentation can be decoupled with the segmentation of the whole eye-region (the combination of visible parts of pupil, iris and sclera). 

As shown in Fig.~\ref{fig:decouple}, the visibility of pupil/iris in the image is determined by the status of eye (openness of eyelids, gaze direction, etc.).

Formally, let $R_{e}$ be the eye-region and $R^{f}_{p}$/$R^{v}_{p}$ denote the \textit{full}/\textit{visible} region of pupil respectively. 
For a typical pupil segmentation approach, the optimization target is to maximize the probability $P(x\in{R^{v}_{p}})$ where $x$ is an arbitrary pixel inside pupil.
Given the afore-mentioned prior, it is obvious that $R^{v}_{p} = R^{f}_{p}\cap{R_e}$.
Therefore, the objective can be further factorized as below:
\begin{equation}
  \mathbf{max} \ P(x\in{R^{v}_p}) = P(x\in{R^{f}_p}, x\in{R_e}) =P(x\in{R^{f}_p}|x\in{R_e})P(x\in{R_e})
\label{eq:prob}
\end{equation}
where $P(x\in{R_e})$ can be implemented through a segmentation head for the eye-region 
and we can represent $P(x\in{R^{f}_p}|x\in{R_e})$ using a full-pupil predictor plus an intersection operation with the eye mask predicted above.
The same principle can be applied to iris prediction.

\begin{figure*}[htbp]
    \centering
    \includegraphics[height=5.5cm]{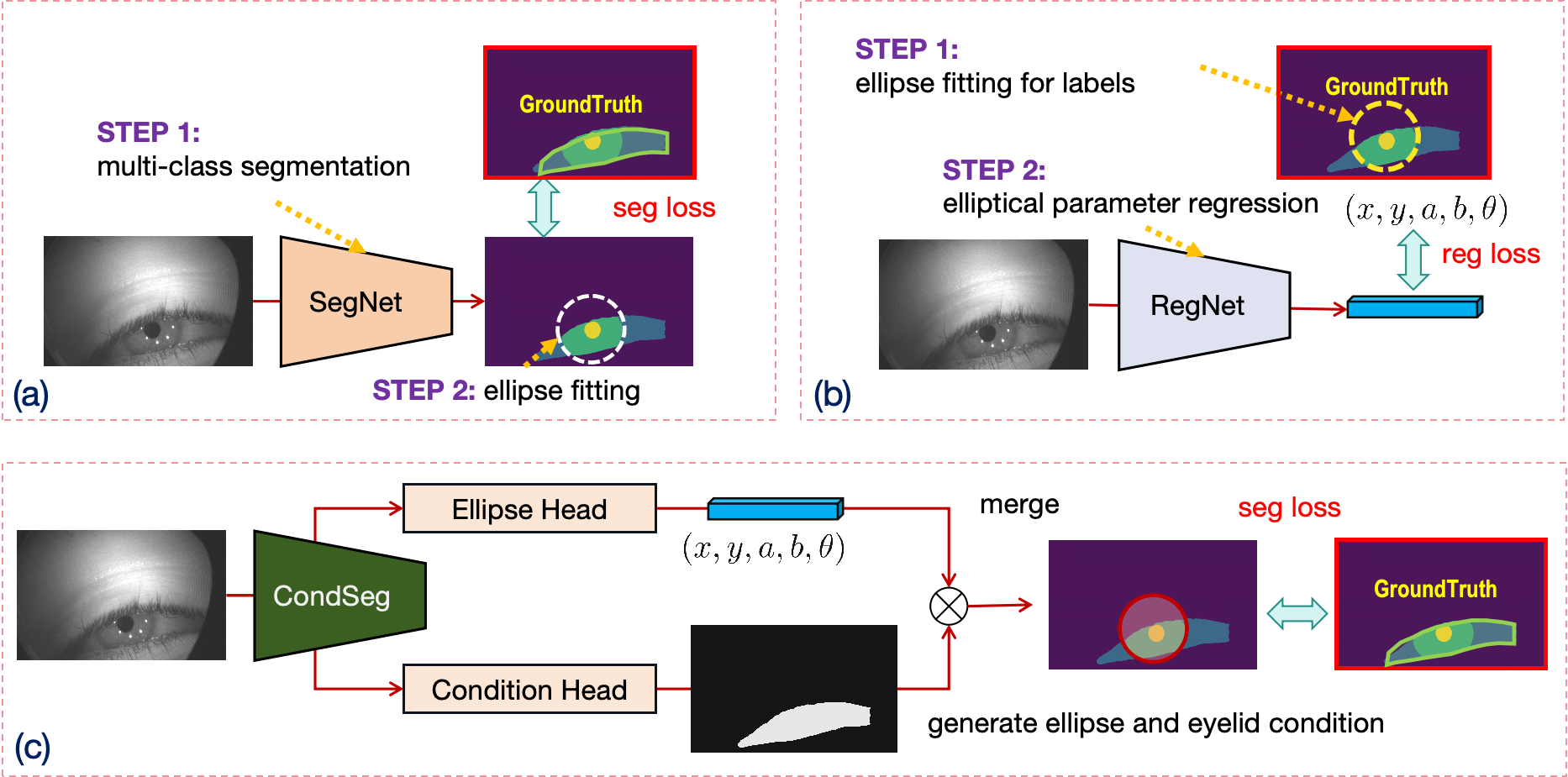}
    \caption{Comparison of different schemes for full ellipse estimation for pupil/iris, where (a) trains a multi-class segmentation first and uses the predicted dense mask to fit ellipse parameters as post-processing, and (b) first generates elliptical parameters for each sample as ground-truth, and trains a regression network to predict the parameters. Our proposed strategy is illustrated in (c), which directly predict the elliptical parameters without explicit ellipse annotations.}
    \label{fig:pipeline}
\end{figure*}

Based on the formulations above, the pipeline of our method is shown in Fig. \ref{fig:pipeline}. Our network (denoted as \textcolor{RoyalPurple}{\textbf{CondSeg}}) produces two components: the eye-region mask and ellipses of full pupil/iris in 5D parameter format, then the predicted full pupil/iris region is merged with eye-region mask to generate the eyelid-occluded pupil/iris, which makes it possible to calculate loss to train the network. In this pipeline, we require no explicit 5D elliptical parameter as regression target, and no post-processing procedure to fit ellipse for the output segmentation masks either, therefore reducing the cost of manual annotations and pre-/post-processing.

A crucial step in the proposed approach is the conversion from 5D elliptical parameter to segmentation mask, which must keep all calculations differentiable to optimize the network parameters. To solve this problem, we transform the 5D parameters to general elliptical equation in matrix form $\bm{x}^\mathrm{T} \bm{M} \bm{x} = 0$, and calculate the value for each coordinate to obtain the distance map. The sign of value for one position in the distance map can indicate this position is inside or outside of the ellipse. Then the distance map is transformed to a soft segmentation map and the points inside the eye-region mask are used to measure the correctness of ellipse prediction. For inference stage, the CondSeg network produces both eye-region mask and pupil/iris ellipses directly, and the commonly used segmentation mask can be generated by simply multiplying the eye-region mask with full pupil/iris masks.

To summarize, the main contributions of our paper are as follows:

\begin{itemize}

\item[1.] We use the prior knowledge by analyzing the relations of classes in eye segmentation tasks, and transform the problem from multi-class segmentation to conditioned segmentation implemented by decoupled prediction of eye-region and pupil/iris ellipses.

\item[2.] To directly encode the elliptical prior into the model, we introduced an approach to transform the 5D elliptical parameter to soft segmentation mask, which allows network optimization in pixel-wise manner as in segmentation tasks for elliptical parameters (instead of in regression manner, which need ground-truth parameters).

\item[3.] The proposed pipeline for ellipse estimation of full pupil/iris is  simple yet effective, where no explicit elliptical parameter annotations are required, thus reducing the annotation burden.
\end{itemize}


\section{Related Works}

As for the importance of eye parsing in real world AR/VR areas, many efforts are made to deal with this task. As the location of pupil is the most essential factor in estimating gaze, designing robust and accurate pupil detection methods has been well studied. ElSe \cite{fuhl2016else} applies Canny edge filtered image to evaluate and select best fitting ellipse for pupil detection. ExCuSe \cite{fuhl2015excuse} uses edge filtering and oriented histograms to find the pupil location. PuRe \cite{santini2018pure} also works in edge map using edge segment selection and combination, which can process in real-time and produce confidence measure for candidate pupil.

Segmentation of iris and sclera is also of vital importance for their possible usage as bio-metric feature in recognition area \cite{alkassar2015robust, kerrigan2019iris}. For these two components, CNN-based methods are exploited to extract features implicitly. In \cite{wang2019joint}, a multi-task learning framework is proposed to boost iris segmentation results. Another work \cite{park2018learning} considers the eye feature extraction task as landmark localization, and trains a stacked-hourglass network to predict the landmarks that can be used as input to gaze estimation methods. DeepVOG \cite{yiu2019deepvog} uses FCNN (fully convolutional neural network) to segment pupil region, and conduct gaze estimation based on the fitted contour. In order to focus on the designing of networks and training strategies, the problem of eye parsing is reformulated as multi-class semantic segmentation task. In this setting, RITnet \cite{chaudhary2019ritnet} combines U-Net and DenseNet for designing a real-time eye segmentation network, and trains the model with domain-specific augmentations and boundary related loss functions. Another difficulty for segmenting eye images is the requirement for large manual annotated training dataset, RIT-Eyes \cite{nair2020rit} employs rendering-based method to get synthetic dataset containing various conditions for training segmentation models, while \cite{chaudhary2021semi} explores semi-supervised learning schemes to make full use of unlabelled images to assist training when only a few labelled images can be obtained.

In the above mentioned methods, common strategy for estimating full pupil/iris ellipses is to segment the visible regions firstly and use the predicted partial masks to fit ellipse for future use \cite{yiu2019deepvog}, as in Fig.\ref{fig:pipeline}(a), another way is to use pre-annotated full pupil/iris labels instead of partial annotations (only visible part inside eye-region) \cite{park2018learning, wang2019joint}, then the network is trained with annotated ground-truth to produce dense masks with full pupil/iris information (area, landmark, or boundary), as in Fig.\ref{fig:pipeline}(b). EllSeg \cite{kothari2021ellseg} aims at obtaining full ellipses of pupil/iris by pixel-wise predicting with full ellipse masks. The reason why it works is that segmentation networks can map the pixel to the correct category even in occluded region due to spatial correlations and context. Though EllSeg achieves satisfactory results, it has two deficiencies: one is the need for generating full ellipse masks and parameters, and the other is the segmentation mask may not be strictly elliptical, analytic form of ellipse from regression is not always compatible with the segmentation mask.

To encode the ellipse prior explicitly into the model design, we refer to the general ellipse detection based method \cite{wang2022eldet} which regresses 5D elliptic parameters via minimizing the difference of predicted and ground-truth elliptical parameter, with mask segmentation as auxiliary. Direct prediction of 5D parameters can constrain the output mask to be exact elliptical, but still need regression target. In our work, we abandon the parameter regression scheme to optimize 5D elliptical parameters, and use segmentation annotations for supervision.

\section{Methodology}
\subsection{Overall Pipeline and Network Architecture}


\begin{figure*}[ht]
    \centering
    \includegraphics[width=0.95\linewidth]{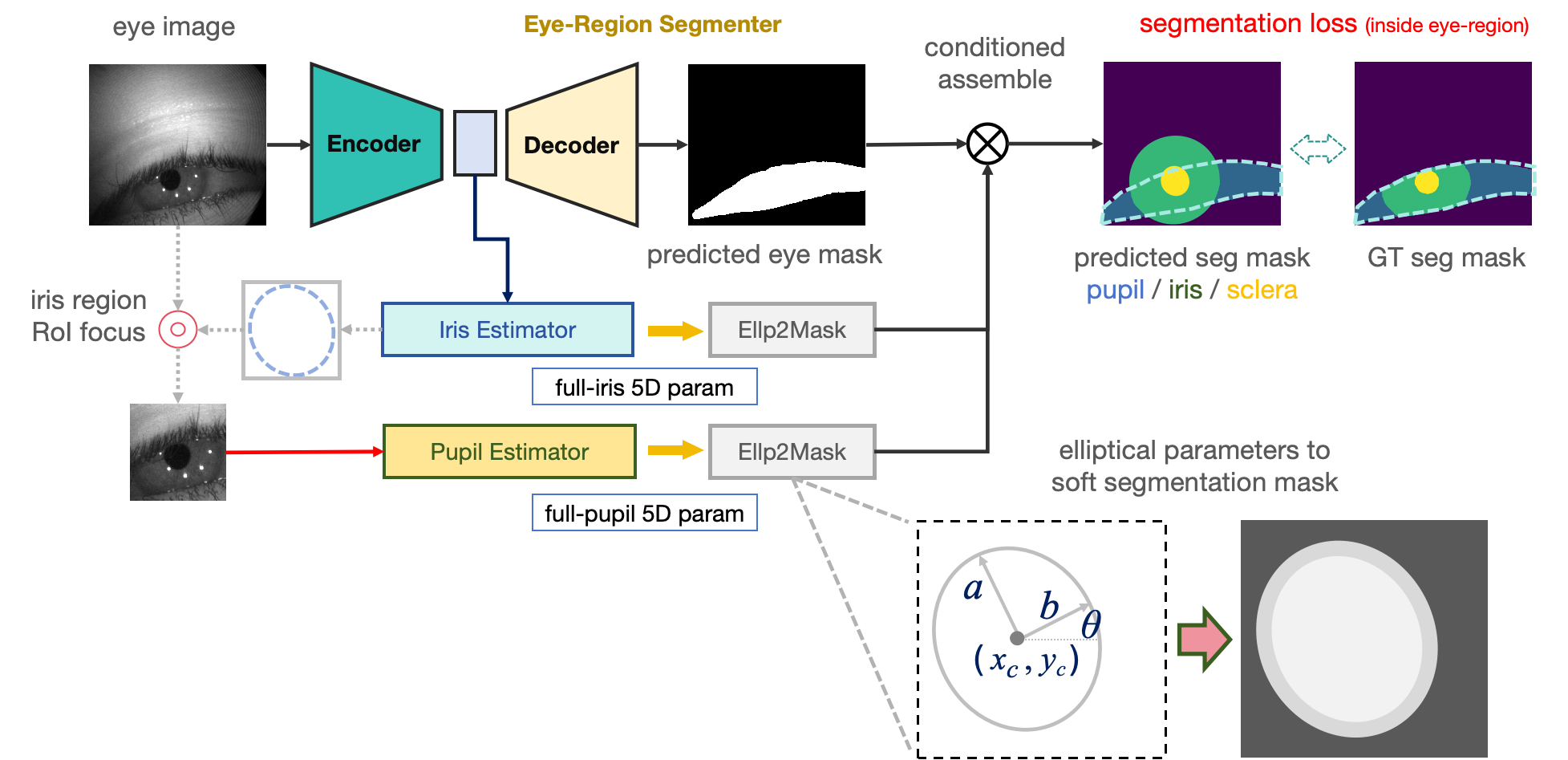}
    \caption{Network architecture of proposed method. Dense-block based encoder-decoder network extracts image features and predicts eye-region segmentation mask. The encoded feature is also utilized to estimate iris elliptical parameters. Full-pupil ellipse is predicted from the cropped full iris RoI region. All elliptical parameters are converted to soft segmentation mask (conditioned by eye-region) for calculating loss to optimize the correctness of elliptical parameters.}
    \label{fig:network}
\end{figure*}

The network architecture of proposed conditioned segmentation method is shown in Fig. \ref{fig:network}. Firstly, a dense-block based encoder \cite{chaudhary2019ritnet} serves as a backbone to extract features from given eye image, which is used both for iris elliptical parameter prediction and eye-region segmentation. For the pathway of \textcolor{RedOrange}{\textbf{Eye-Region Segmenter}}, a decoder outputs a pixel-wise eye-region segmentation map. The eye-region map is applied as condition of pupil and iris as stated before, controlling the visible and invisible parts of predicted pupil and iris. Another pathway is the \textcolor{RoyalPurple}{\textbf{Iris Estimator}}, using the encoded features to estimate the elliptical parameters of full-iris with multi-MLP layers. The prior knowledge about eye image tells us that the full-pupil usually lies within the full-iris region, so for precise estimation of pupil ellipse, the bounding box of full-iris ellipse is used to crop the eye image to force the \textcolor{BrickRed}{\textbf{Pupil Estimator}} to focus on the iris region, which reduces the difficulty of pupil parameter estimation by leveraging relations of different categories. The structure of pupil estimator is similar to iris estimator, which only differs in the number of layers and feature channels.

The key step of proposed CondSeg is to train the network without explicit elliptical parameter as ground-truth. To deal with this problem, we design a differentiable module, denoted as \textcolor{Magenta}{\textbf{Ellp2Mask}} in Fig.\ref{fig:network}, which can convert the 5D elliptical parameter to soft segmentation mask. The detailed calculation will be illustrated in the following contents. After the pupil/iris segmentation map is obtained, then eye-region mask is used as an ignorance mask, where only pixels inside the eye-region calculate loss and back-propagate gradients to optimize the network parameters, while the ones outside eye-region are regarded as ``ignored''. Finally, when the network parameters converge, the pipeline can produce both full-iris and full-pupil ellipses directly, as well as a 3-class segmentation map by conditioned assemble for parsing pupil/iris/sclera regions.


\subsection{Estimators for Full Iris and Pupil Ellipses}

For compatibility across different data sources, the 5D elliptical parameters generated by pupil/iris estimator is set to be the relative value based on the input size. The output of MLP is normalized by Sigmoid activation to constrain the values within $(0, 1)$. Moreover, to avoid extreme $a$ and $b$ values in 5D parameters, a minimum value $\varepsilon$ is added to the predicted axis lengths. The absolute values are converted from estimator predictions via the following formula (variables with ``\^{}'' refer to the direct output of estimator, ``$_{(i)}$'' is for ``iris'', and $h$, $w$ are height and width of the image respectively):


\begin{equation}
    \begin{aligned}
    & x_{0(i)} = \hat{x}_{0(i)} * w, \quad y_{0(i)} = \hat{y}_{0(i)} * h, \quad \theta_{(i)} = \hat{\theta}_{(i)} 
    \\ a_{(i)} = & (\hat{a}_{(i)} + \varepsilon) * \min(w, h) / 2, \quad
    b_{(i)} = (\hat{b}_{(i)} + \varepsilon) * \min(w, h) / 2\\
    \end{aligned}
\end{equation}

In order to maintain the ellipse shape of iris and pupil, aspect ratio of each image is preserved in the training process where input image is resized. For pupil estimation, we use the minimum bounding square (of iris ellipse) instead of rectangle for cropping the RoI region, then the squares are resized to the same size to train the pupil estimator. The bounding square parameters $(x_1, y_1, s)$ with top-left corner $(x_1, y_1)$ and side length $s$ can be directly calculated from the elliptical parameters:


\begin{equation}
    \begin{aligned}
    \Delta w = \sqrt{a_{(i)}^2\cos^2\theta_{(i)} + b_{(i)}^2\sin^2\theta_{(i)}}, \quad &
    \Delta h = \sqrt{a_{(i)}^2\sin^2\theta_{(i)} + b_{(i)}^2\cos^2\theta_{(i)}} \\
    x_1 = x_{0(i)} - \Delta w, \quad y_1 = y_{0(i)} - \Delta h, & \quad
    s = 2 * \max(\Delta w, \Delta h) \\
    \end{aligned}
\end{equation}

In the inference phase, eye components are estimated via combined assemble, where iris elliptical parameters and eye-region mask are predicted and converted according to image size first, then iris RoI region is cropped for estimating pupil parameters. After elliptical parameters for pupil are obtained, the axis lengths $a$ and $b$ are converted according to the cropped square size, and the ellipse center is translated to the original position according to the iris bounding square (``$_{(p)}$'' is for ``pupil''):


\begin{equation}
    \begin{aligned}
    & x_{0(p)} = \hat{x}_{0(p)} * s + x_1, \quad y_{0(p)} = \hat{y}_{0(p)} * s + y_1, \quad \theta_{(p)} = \hat{\theta}_{(p)} 
    \\ & a_{(p)} = (\hat{a}_{(p)} + \varepsilon) * s / 2, \quad
    b_{(p)} = (\hat{b}_{(p)} + \varepsilon) * s / 2\\
    \end{aligned}
\end{equation}


\subsection{From Elliptical Parameter to Segmentation Mask}

In this subsection, we will illustrate the process which can convert elliptical parameters towards segmentation mask in a differentiable way. Firstly, we denote the predicted elliptical parameter as $(x_0, y_0, a, b, \theta)$, $(x_0, y_0)$ is the center of ellipse, $a$ and $b$ are semi-major and semi-minor axis length, and $\theta$ is the angle between semi-major axis and $x$-axis. Using the standard ellipse equation and considering shift and rotation, the ellipse equation is:

\begin{equation}
    \frac{(\hat{x}\cos\theta + \hat{y}\sin\theta)^2}{a^2} + \frac{(-\hat{x}\sin\theta + \hat{y}\cos\theta)^2}{b^2} = 1, \ \text{where} \ \hat{x} = x - x_0, \ \hat{y} = y - y_0
\end{equation}

Moreover, considering conic section (including ellipse) can be represented via quadratic equation in two variables in Cartesian coordinate system, the general form of which is:


\begin{equation}
    Ax^2 + Bxy + Cy^2 + Dx + Ey + F = 0, \quad \text{where } A, B, C \neq 0    
\end{equation}

Comparing the standard equation and the general form equation, the parameters follow the relations as:

\begin{equation}
    \begin{aligned}
    &A = \sin^2\theta / b^2 + \cos^2\theta / a^2, &
    &B = 2(1/a^2 - 1/b^2)\sin\theta\cos\theta \\
    &C = \cos^2\theta / b^2 + \sin^2\theta / a^2, &
    &D = -2Ax_0 - By_0 \\
    &E = -Bx_0 - 2Cy_0, &
    &F = -(Dx_0 + Ey_0) / 2 - 1
    \end{aligned}
\end{equation}

The above general form equation can also be written in matrix notation as:

where the $\bm{x}$ is the augmented coordinate and $\bm{M}$ the ellipse matrix:

\begin{equation}
    \begin{split}
    \bm{x}^{\mathrm{T}} \bm{M} \bm{x} = 0 \text{,} \quad \text{ where }  
    \bm{x} = [x, y, 1]^{\mathrm{T}} &\text{,} \quad \bm{M} = \begin{bmatrix}
     A & B/2 & D/2\\
     B/2 & C & E/2\\
     D/2 & E/2 & F
    \end{bmatrix}
    \end{split}
\end{equation}

\begin{figure*}[h]
    \centering
    \includegraphics[width=0.7\linewidth]{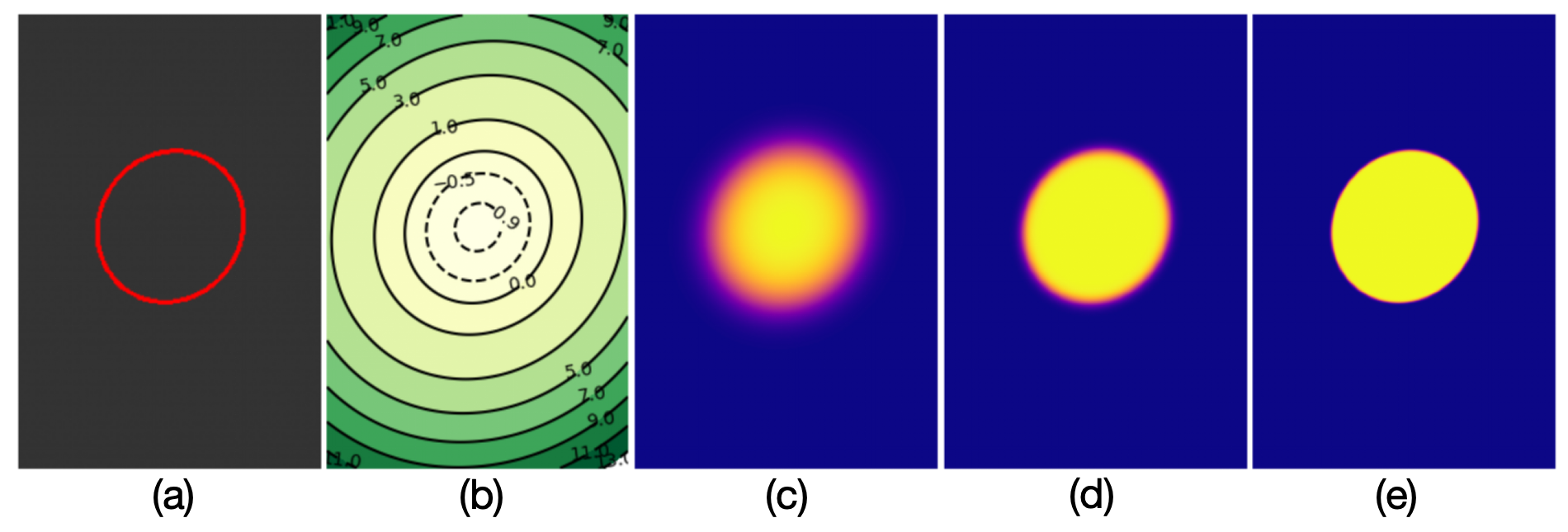}
    \caption{Ellipse drawn directly from parameters is shown in (a), and (b) is the \textit{distmap} $\bm{D}$ which is calculated with $\bm{x}^{\mathrm{T}} \bm{M} \bm{x}$, (c)-(e) are \textit{segmap}s with $\tau=50, 200$ and $1000$}
    \label{fig:distmap}
\vspace{-10pt}
\end{figure*}

As the ellipse corresponds to the equation $\bm{x}^{\mathrm{T}} \bm{M} \bm{x} = 0$, then the two inequalities $\bm{x}^{\mathrm{T}} \bm{M} \bm{x} > 0$ and $\bm{x}^{\mathrm{T}} \bm{M} \bm{x} < 0$ corresponds to the outside and inside of the given ellipse respectively. After the elliptical parameter $(x_0, y_0, a, b, \theta)$ is predicted by pupil/iris estimation heads, we can generate a map with the same size as the ground-truth segmentation map, and use the matrix converted from elliptical parameter and coordinate of each pixel to calculate the value of $\bm{x}^{\mathrm{T}} \bm{M} \bm{x}$, which results in the denoted \textit{distmap} $\bm{D}$. The range of values in $\bm{D}$ is $[-1, +\infty)$ (without image size restrictions), which cannot be used as segmentation map directly. The following process is used to convert $\bm{D}$ into \textit{segmap} $\bm{S}$:

$$
\bm{S} = \sigma(\frac{-\bm{D}}{\max(\bm{D})+\delta} * \tau), \quad \bm{D} = \bm{x}^{\mathrm{T}} \bm{M} \bm{x}
$$

where $\sigma$ refers to Sigmoid function to map $(-\infty, +\infty)$ to $(0, 1)$, and $\tau$ is the hyper-parameter controlling the smoothness of transition area. Multiplying $\bm{S}$ with eye-region mask gives the predicted part segmentation mask of pupil/iris, binary cross entropy loss is then conducted with ground-truth segmentation mask for optimization. Fig. \ref{fig:distmap} is an example for visualization of the ellipse, \textit{distmap}, and \textit{segmap} with different $\tau$ values.


\section{Experiments}

\subsection{Datasets and Evaluation Metrics}

To demonstrate the effectiveness of our proposed CondSeg method, we conduct experiments on two widely used public datasets: OpenEDS-2019\cite{garbin2019openeds} and OpenEDS2020\cite{palmero2020openeds2020}. Both datasets are collected on head-mounted VR display devices with eye-facing cameras. OpenEDS-2019 includes totally 152 subjects with different genders and ages, which is split into balanced train/validation/test sets. Its successor OpenEDS-2020 is collected from 74 subjects and contains 200 selected video sequences, its testset is acquired by selecting 5 frames in each sequence. As our task is for full pupil and iris estimation, we discarded the samples without valid visible pupil region. Moreover, we used the setting from EllSeg which crops the image to keep only the surroundings of eye-region, resulting in different image sizes from the original datasets. Detailed information of OpenEDS-2019/-2020 datasets is shown in Tab.\ref{tab:dataset}

\subsection{Implementation Details}

We train proposed CondSeg in a two-stage strategy, where the iris elliptical parameter and eye-region mask is trained first to get a robust and precise full iris region to serve as prior for full pupil estimation. Then the pupil estimator is trained to fit the visible pupil region. After both stages are finished, we can apply CondSeg to infer on samples of testset to validate the performance of the method.

\begin{table}[]
\caption{\small{
Dataset information for training and evaluation. The \textit{no. sub.} refers to the number of subjects included in the data, and \textit{image size} is the size of cropped images with a similar method to EllSeg. Invalid samples with no visible pupils are filtered in our experiments.
}}
\centering
\begin{tabular}{c|c|c|c|c}
\toprule
dataset & no. sub.&  train samples & test samples & image size ($w\times h$) \\ \hline
OpenEDS-2019 & 152 &  8887     &  3836   &  400 $\times$ 300 \\
OpenEDS-2020 & 74 &  1548     &  960  & 640 $\times$ 300 \\
\bottomrule
\end{tabular}
\label{tab:dataset}
\end{table}


The parameters and settings for our experiments are listed as below. The first training stage is done on 8 GPUs with 8 samples per batch (on each GPU) for OpenEDS-2019, and 2 samples per batch for OpenEDS-2020. The input size is $320 \times 240)$ (OpenEDS-2019) and $512 \times 240)$ (OpenEDS-2020) respectively. The first stage is trained for 200 epoches for OpenEDS-2019 and 300 for OpenEDS-2020. For training pupil in RoI region, the batch size of each GPU is set to 4 in both datasets, and input size of cropped square is $200\times200$. Training epoches is 150 (OpenEDS-2019) and 500 (OpenEDS-2020). For both stage, we use AdamW optimizer with initial learning rate 0.0004, the learning rate is reduced by 0.2 in 1/6, 1/3, 1/2 and 5/6 of total epoch number using MultiStepLR scheduler. The minimum value of relative axis length, i.e. $\delta$, is set to 0.01 for iris and 0.1 for pupil, and hyper-parameter $\tau$ is set to 800 in all experiments. Augmentations including random flip, rotate, noise injection, blur, and luminance adjustment are used in both training stages to enhance the generalization ability.

\subsection{Comparison of Eye Parsing on Visible Parts}

The IoU (intersection-over-union) of each category between predicted and annotated segmentation masks (including visible pupil, iris and sclera) provides a effective metric to validate the fitting performance of eye parsing methods when ground-truth full elliptical masks are not available. As our CondSeg does not require annotated full ellipse masks, we first compare the IoUs of each part within eye-region for different methods. 

The method denoted as PartSeg-baseline is the common multi-class semantic segmentation model with the same network backbone as encoder-decoder part of CondSeg (except for the output channel number). As the PartSeg model only gives visible masks of pupil and iris, ellipse fitting with mask edge points as post-processing step is essential if we want to obtain the full pupil and iris. EllSeg-Seg and -Ellp is the segmentation and regression results from EllSeg model. Both results are from the same EllSeg model which is trained using pre-computed (by RANSAC\cite{fischler1981random}) ellipses from semantic masks, the EllSeg-Seg path learns the segmentation mask of full pupil and iris in a pixel-wise manner, while EllSeg-Ellp regresses the ground-truth 5D elliptical parameters directly. Because of the EllSeg model provides no eye-region mask, we use ground-truth eye-region mask to calculate the inside IoUs.

\begin{table}[htbp]
\caption{\small{
{Results for eye parsing task (in metrics of IoU with ground-truth masks). EllSeg-Seg and -Ellp is the outputs of two pathways of EllSeg, $\dagger$ means condition of eyelid is from ground-truth eye-region masks to make fair comparison with EllSeg (only predict pupil/iris, shown results are also with ground-truth eye-region masks). Column with ``elli. GT'' refers to if the method requires prepared explicit elliptical parameters, and ``post-fit elli.'' means the method still needs to fit ellipse as post-processing.}
}}
\scriptsize
\centering
\begin{tabular}{l|ccc|ccc|c|c}
\toprule
\makecell{\textbf{Method}}  & \multicolumn{3}{c|}{\textbf{OpenEDS-2019}} & \multicolumn{3}{c|}{\textbf{OpenEDS-2020}} & \multicolumn{2}{c}{}\\ \hline
& pupil   & iris-region   & eye-region   & pupil   & iris-region   & eye-region  & elli. GT & post-fit elli. \\ \hline 
PartSeg-baseline     &  91.34  &  94.06 &  94.87 &  92.18 &  94.27  &  95.76  & \XSolidBrush & \CheckmarkBold \\
EllSeg-Seg           &  93.07  &  97.37 &  -  &  91.28  &  96.18  &  -  & \CheckmarkBold & \CheckmarkBold  \\
EllSeg-Ellp          &  90.69  &  94.59 &  -  &  86.75  &  92.50  &  -  & \CheckmarkBold & \XSolidBrush  \\
\textbf{CondSeg$^{\dagger}$}  &  91.11  &  95.86 &  -  &  87.08  &  94.79 &  -  & \XSolidBrush & \XSolidBrush \\ 
\textbf{CondSeg}              &  90.91  &  94.37 & 95.94 &  86.80 & 91.83 & 90.51  & \XSolidBrush & \XSolidBrush \\ 
\bottomrule
\end{tabular}
\label{tab:ious}
\end{table}

Note that we compare iris-region (visible regions inside iris ellipse, including pupil) instead of iris mask (only iris part inside eye-region, without pupil) to decouple the influence of performance of pupil from iris, which is more reasonable as the iris and pupil are processed separately. The EllSeg model is tested for OpenEDS-2019 with weights pre-trained on OpenEDS-2019 dataset in the official codebase. For OpenEDS-2020, there is no official pre-trained weights on OpenEDS-2020, so we train EllSeg by ourselves with the same dataset setting as CondSeg on OpenEDS-2020 using the official training code.

\begin{figure*}[htbp]
    \centering
    \includegraphics[width=0.75\linewidth]{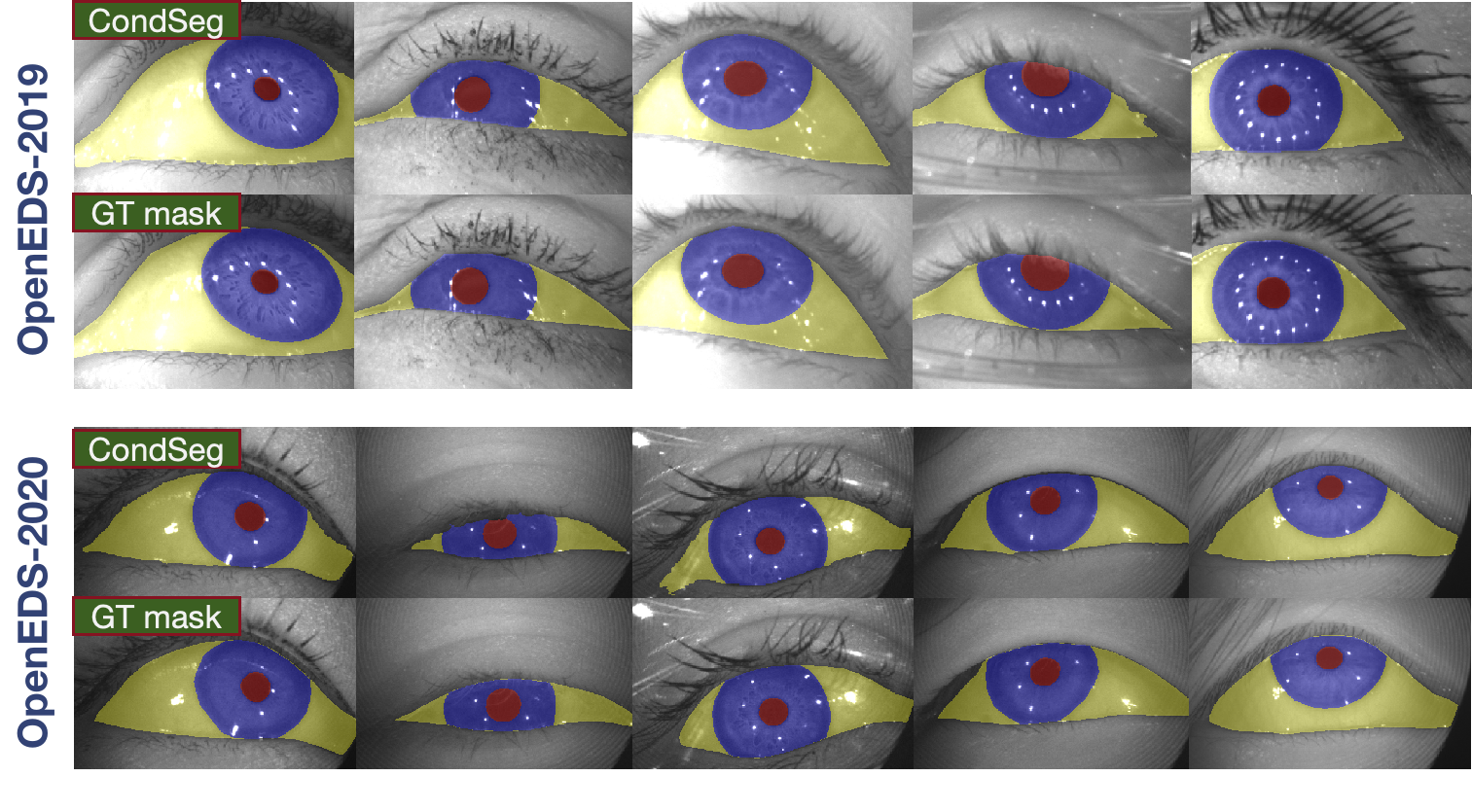}
    \caption{Eye-parsing performances on tested public datasets OpenEDS-2019 and OpenEDS-2020 is shown above by comparing the output of CondSeg and ground-truth segmentation masks. Note that CondSeg can still provide reasonable pupil and iris masks even when they are obviously occluded by eyelid.}
    \label{fig:segout}
\end{figure*}

The last two rows show performance of CondSeg. Our CondSeg model outputs full pupil/iris and eye-region mask, which can be assembled as the semantic mask format. To make a fair comparison with EllSeg model, we calculated additionally CondSeg$^\dagger$ using ground-truth eye-region masks. From Tab.\ref{tab:ious}, the IoUs of pupil and iris-region of CondSeg is competitive or even better than EllSeg-Ellp, indicating that our conditioned learning strategy with partial visible masks is as effective as regressing on ground-truth explicit elliptical parameters, but in a more efficient and elegant manner (without preparing ground-truth ellipses). The setting of EllSeg-Seg is more accurate in predicting visible masks as shown in Tab.\ref{tab:ious}, this advantage is reasonable as the EllSeg-Seg setting is a pixel-wise semantic segmentation task. Some samples of results produced by CondSeg are shown in Fig. \ref{fig:segout}. However, EllSeg-Seg is not robust enough to comply with the ellipse prior, especially in the heavily occulded cases. EllSeg-Seg may fail in expanding the invisible parts of pupil and iris as ellipses, which hinders the ellipse fitting post-processing using predicted full elliptical masks. 

\vspace{-15pt}

\subsection{Analysis of Full Ellipses for Full Pupil and Iris}

\vspace{-15pt}

\begin{table}[htbp]
\caption{\small{
{Full pupil center location error and iris center location error (in parenthesis). Errors are measured in pixel. \textit{ellipse fitting} refers to wether post-processing fitting is necessary for finding the pupil and iris center. EllSeg-Seg provides ellipse center by fitting ellipse using full mask contour, while EllSeg-Ellp and CondSeg can directly output center coordinates.}
}}
\centering
\begin{tabular}{l|c|c|c}
\toprule
& OpenEDS-2019 & OpenEDS-2020 & ellipse fitting \\ \hline 
PartSeg-baseline &  1.52 (3.83)  &  1.87 (5.43)  & RANSAC \\
EllSeg-Seg   &  1.47 (2.38) &  1.35 (4.34) & contour \\
EllSeg-Ellp  &  1.12 (2.10) &  0.88 (4.81) & \XSolidBrush \\
CondSeg        &  1.48 (3.42) &  1.61 (5.91) & \XSolidBrush \\ 
\bottomrule
\end{tabular}
\vskip 5pt plus 1fil
\label{tab:errors}
\end{table}

\vspace{-5pt}

In order to demonstrate the performance of full pupil/iris estimation, it is required to calculate the location accuracy of full pupil/iris in test datasets. Pupil and iris location error is a suitable metric for evaluating the location accuracy of full ellipses. However, as our setting implies, their is no full elliptical parameters as ground-truth. To deal with the lack of ground-truth labels, we use RANSAC to fit the full ellipses with part segmentation masks of iris and pupil, and select the top-$k$ ($k=500$) most precise samples as ground-truth full ellipse labels (samples which cannot be fitted well are ignored because it may not give the real locations of pupil and iris centers). We show the median value of location errors of pupil and iris from all tested samples in Tab.\ref{tab:errors} inspired by evaluations of EllSeg experiments.

\begin{figure*}[htp]
    \centering
    \includegraphics[width=0.75\linewidth]{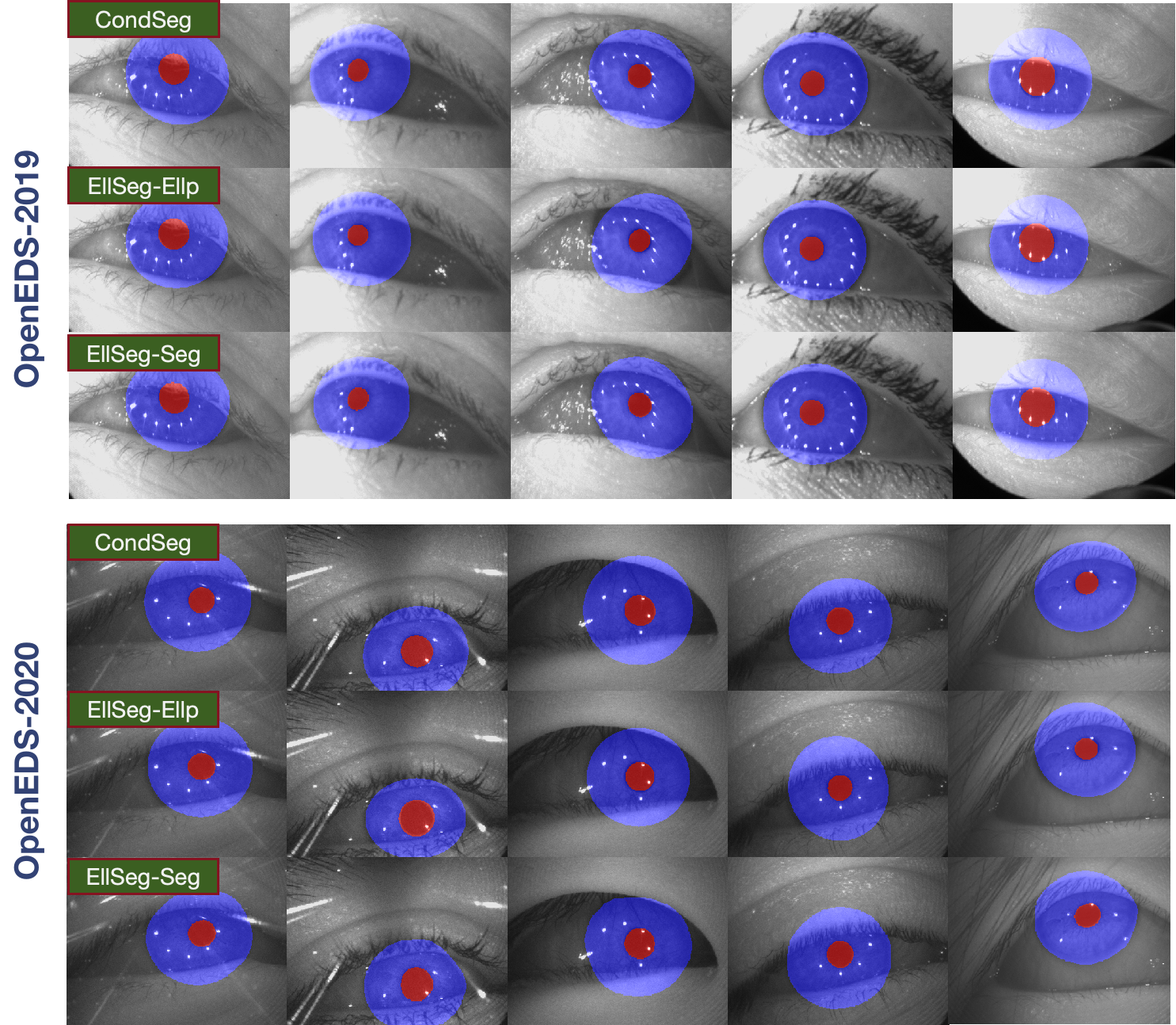}
    \caption{Full pupil and iris estimation results of CondSeg, comparing with EllSeg (with both modals denoted as EllSeg-Ellp and EllSeg-Seg) trained on OpenEDS-2019 and OpenEDS-2020 respectively. Qualitative performance validated the effectiveness of our CondSeg model, implying the potential to estimate the full elliptical pupil and iris region without ground-truth full masks or explicit parameter annotations.}
    \label{fig:ellout}
\end{figure*}

\vspace{-3pt}

In Tab.\ref{tab:errors}, the row with ``PartSeg-baseline'' shows the fitted ellipse center error using the common segmentation model and post-processing RANSAC fitting. EllSeg-Seg is expected to output full iris and pupil mask, which can be directly used to fit ellipses or calculate pupil/iris center. 
EllSeg-Ellp and CondSeg output ellipse center as a parameter, so the location error is calculated by simply comparing the ground-truth pupil/iris center with predicted ones. Results show that CondSeg has the capability to generate relatively accurate estimations for full pupil and iris without pre-annotated training targets. The elliptical prior is valid in conditioned training processes to identify the visible part of pupil/iris belongs to which part of full ellipse, leading to robust and correct locations. Fig. \ref{fig:ellout} shows some samples for full pupil and iris estimation results from CondSeg, EllSeg-Seg and -Ellp. Our CondSeg can produce competitive results with EllSeg-Ellp, yet without any full pupil and iris masks.


\subsection{Iris Region RoI Focus and Augmentations}

Cropping iris region as RoI is beneficial for a better and robust training of pupil estimator. This processing encompasses the prior of pupil and iris location, and makes the model to focus on their relative location and scale of full pupil, therefore reduces the difficulty for pupil estimation, and cancelled the barrier to locate pupil in the whole image. Moreover, as our loss is calculated on \textit{segmap}, augmentations for segmentation task (flip/rorate/blur/noise etc.) can also improve the training performance of CondSeg model. After the Ellp2Mask module, our problem is re-formulated as binary semantic segmentation, where the augmentations preserving elliptical prior all can be utilized to enhance the performance. Tab.\ref{tab:ablation} shows the ablation for iris-region RoI focus and augmentation on OpenEDS-2019. The top row is conducted by directly outputing two 5D parameters simultaneously for full pupil and iris. Comparison of different settings indicates the effectiveness of both iris-region RoI focus and augmentations.

\vspace{-0.5cm}
\begin{table}[hb]
\caption{\small{Ablation Study for iris-region RoI focus (denoted as \textit{iris foc.}) and augmentations in the training process (denoted as \textit{aug.}). \textit{IoU$_{p}$} is for pupil IoU inside eye-region, and \textit{err-loc$_p$} and \textit{err-loc$_i$} denote center location errors of pupil and iris.}}
\centering
\begin{tabular}{c|c|c|c|c}
\toprule
iris foc. & aug. &  IoU$_{p}$ & err-loc$_{p}$ & err-loc$_{i}$  \\ \hline
         &      &  86.78    &  1.98  &  3.92\\
\CheckmarkBold &    &  90.01     &  1.63  &  3.89\\
\CheckmarkBold & \CheckmarkBold & 91.11     &  1.48  &  3.42\\
\bottomrule
\end{tabular}
\vskip 5pt plus 1fil
\label{tab:ablation}
\end{table}

\vspace{-30pt}
\section{Conclusion}

\vspace{-5pt}

In this paper, we propose a novel approach \textcolor{RoyalPurple}{\textbf{CondSeg}} for full pupil and iris estimation, which is capable of generating 5D elliptical parameters using only common semantic segmentation masks, instead of delicately annotated ellipses. In the view of problem modeling, we decouple the prediction of eye-region and pupil/iris ellipses estimation, and consider eye-region as condition for the full pupil/iris ellipses. Elliptical parameters are converted to soft segmentation masks, and optimized within eye-region condition only. Experiments have validated the robustness and accuracy of CondSeg, which requires no pre- or post-processing, boosting the efficiency for AR/VR related eye-tracking development.

\vspace{-5pt}

%
%
\bibliographystyle{splncs04}
\footnotesize
\bibliography{main}

\end{document}